\documentclass[10pt,twocolumn,letterpaper]{article}

\usepackage[pagenumbers]{wacv} 

\usepackage{graphicx}
\usepackage{amsmath}
\usepackage{amssymb}
\usepackage{booktabs}
\usepackage{multirow}
\usepackage{xspace}
\usepackage{makecell}
\usepackage{pifont}
\usepackage{xcolor,colortbl}         
\usepackage{standalone}
\usepackage{booktabs}
\usepackage{array}
\usepackage{caption}

\usepackage[pagebackref,breaklinks,colorlinks]{hyperref}

 \usepackage{textcomp}
\usepackage{pgfplots}
\usetikzlibrary{patterns,positioning,arrows,arrows.meta,calc,shapes,pgfplots.groupplots,fit,backgrounds}
 \usepackage{textcomp}
 
\usepackage[capitalize]{cleveref}
\crefname{section}{Sec.}{Secs.}
\Crefname{section}{Section}{Sections}
\Crefname{table}{Table}{Tables}
\crefname{table}{Tab.}{Tabs.}

\usepackage{amsmath,amsfonts,bm}

\def\1{\bm{1}}

\DeclareMathAlphabet{\mathsfit}{\encodingdefault}{\sfdefault}{m}{sl}
\SetMathAlphabet{\mathsfit}{bold}{\encodingdefault}{\sfdefault}{bx}{n}

\newcommand{\normltwo}{L^2}

\newcommand{\cmark}{\ding{51}}
\newcommand{\xmark}{\ding{55}}
\newcommand{\W}{W}

\newcommand{\z}{\boldsymbol{z}}

\newcommand{\probpr}{\mathrm{Pr}}

\newcommand{\btheta}{\boldsymbol{\theta}}

\newcommand{\sce}{\mathrm{e}}
\newcommand{\cribo}{CrIBo\xspace}
\newcommand{\lidar}{LiDAR\xspace}
\newcommand{\mname}{S3PT\xspace}
\definecolor{gray60}{gray}{0.8}
\newcolumntype{g}{>{\columncolor{gray60}}c}
\def\wacvPaperID{2072} 
\def\confName{WACV}
\def\confYear{2025}

\begin{document}

\title{S3PT: Scene Semantics and Structure Guided Clustering \\ to Boost Self-Supervised Pre-Training for Autonomous Driving}

\author{
Maciej K. Wozniak$^{1,5*}$ 
\and
Hariprasath Govindarajan$^{2*}$ 
\and
Marvin Klingner$^{1}$ 
\and
Camille Maurice$^{1}$ 
\and
B Ravi Kiran$^{3}$ 
\and
Senthil Yogamani$^{4}$  \and \quad *co-first authors\\
{\tt\small \{mwozniak, hargov, mklingne, cmaurice, ravkira, syogaman\}@qti.qualcomm.com}\\
{\tt\small $^{1}$Automated Driving, Qualcomm Technologies International GmbH} \\
{\tt\small $^{2}$Arriver Software AB and Linköping University, Sweden} \\
{\tt\small$^{3}$Qualcomm SARL, France \quad $^{4}$Automated Driving, Qualcomm Technologies, Inc} \\
{\tt\small $^{5}$KTH Royal Institute of Technology, Sweden}
}

\maketitle

\begin{abstract}
Recent self-supervised clustering-based pre-training techniques like DINO and \cribo have shown impressive results for downstream detection and segmentation tasks. However, real-world applications such as autonomous driving face challenges with imbalanced object class and size distributions and complex scene geometries. In this paper, we propose \textbf{S3PT} a novel scene semantics and structure guided clustering to provide more scene-consistent objectives for self-supervised training. Specifically, our contributions are threefold:
First, we incorporate semantic distribution consistent clustering to encourage better representation of rare classes such as motorcycles or animals.
Second, we introduce object diversity consistent spatial clustering, to handle imbalanced and diverse object sizes, ranging from large background areas to small objects such as pedestrians and traffic signs.
Third, we propose a depth-guided spatial clustering to regularize learning based on geometric information of the scene, thus further refining region separation on the feature level. Our learned representations significantly improve performance in downstream semantic segmentation and 3D object detection tasks on the nuScenes, nuImages, and Cityscapes datasets and show promising domain translation properties. 
\end{abstract}

\begin{figure}
    \centering
    \includegraphics[width=0.78\linewidth, trim=0 0cm 34cm 1cm, clip]{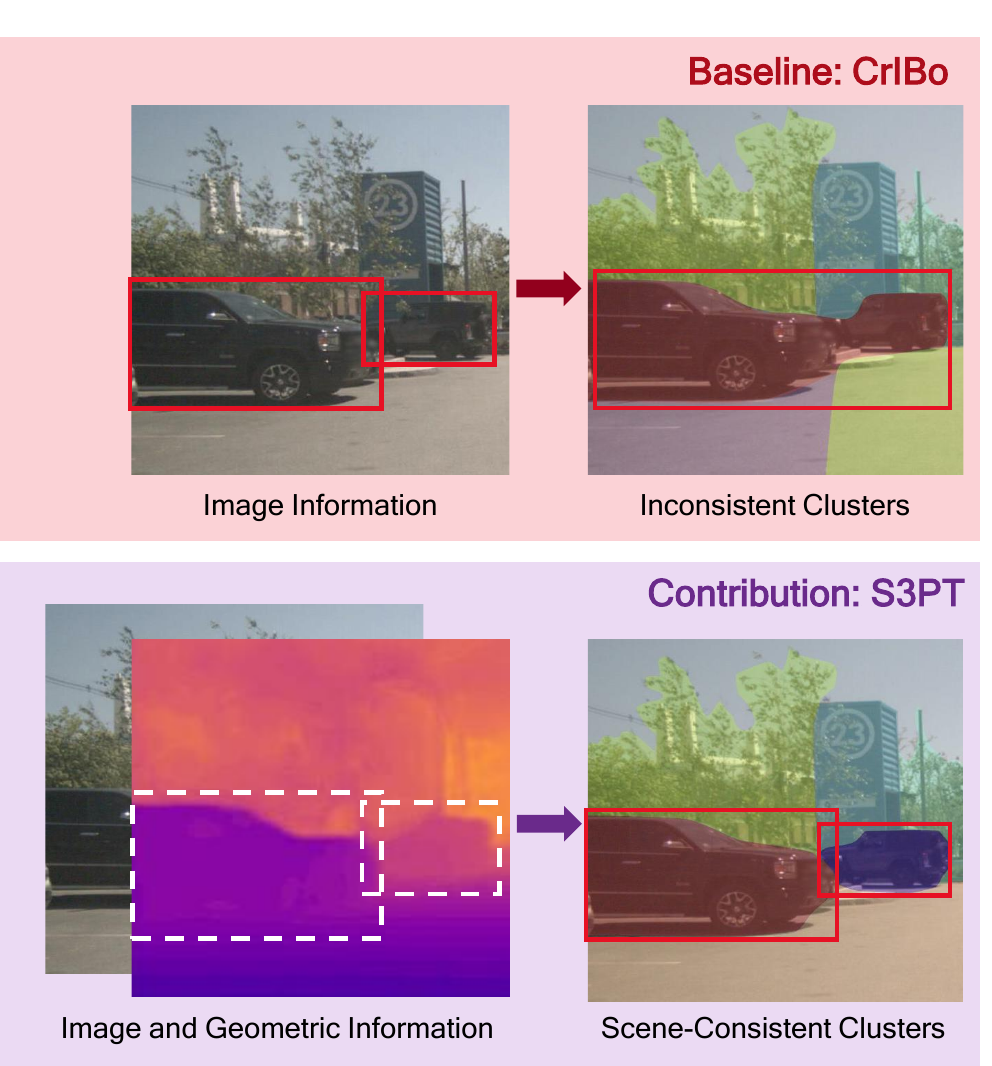}
    \caption{Visualization of our main contribution using scene semantics and structure-guided clustering (S3PT) in self-supervised pre-training. The baseline \cribo (top) provides inconsistent clusters on autonomous driving data , failing to capture geometric information and struggling with imbalanced object sizes and classes due to its strong encouragement of uniform distributions. In contrast, S3PT (bottom) offers scene-consistent clustering (e.g., correctly differentiating the two cars) thanks to our contributions.}
    \label{fig:cribo-ad-highlight}
\end{figure}

\section{Introduction}

Recent research has focused a lot on improving the understanding of vehicle surroundings to increase the safety and autonomy of the vehicles. However, with constant updates to sensor types, setups, and differences between vehicles, it is highly challenging to create one perception system that will achieve high performance regardless of the inference environment, vehicle or sensor setup. While there are many causes, we can think of weather conditions, sensor setup or driving environment that differ during inference and training. This problem is highly relevant for autonomous driving (AD) companies, that iterate between different sensor setups and car models. Another motivation is the challenging scenario when different sensors are available during the training compared to inference. Most modern consumer cars (inference) are equipped with various cameras but do not have \lidar due to design and cost barriers, while data collection vehicles (training) typically have it for accurate 3D data collection and annotation. Leveraging the full data collection sensor suite to create optimal models on the target sensor setup is a problem with many highly relevant practical applications, particularly in autonomous driving.

One promising solution to this problem is self-supervised pre-training, which has proven to be effective at learning informative representations which are transferable to a range of downstream tasks and generalize across different datasets in a similar domain \cite{pretraining_data_ssl}. While these methods easily outperform supervised baselines on image recognition tasks, they are not as strong at dense prediction tasks such as image segmentation and object detection, especially in more challenging settings such as autonomous driving \cite{multisiam}. There is only a limited research focus on developing self-supervised pre-training strategies to learn representations suitable for dense prediction tasks by pre-training directly on complex scene data \cite{croc, cribo, region_cl, vicreg_local, slotcon, odin}. These methods are also evaluated by pre-training on curated datasets such as COCO \cite{dataset_coco}. Training on driving datasets bring up other challenges such as limited diversity (mostly street view, quite repetitive)~\cite{ljungbergh2024neuroncap,tonderski2024neurad,wozniak2024uada3d}. Additionally, those datasets are much smaller than ImageNet or COCO since it is much more costly to collect  the data. Methods such as \cribo assume and encourage uniformly sized object segments and a uniform distribution over the pseudo-labels assigned to these segments. This is in contrast to the distributions typically found in autonomous driving datasets, which do not focus on the caveats presented by self-driving car data, such as small, far away objects, heavy occlusions, and long-tail classes. In general, the applicability of such methods to autonomous driving datasets is not currently known. 

In this work, we build upon the recently published \cribo \cite{cribo}, and propose an improved version, \textbf{\mname}, better adapted to autonomous driving datasets, showed in~\cref{fig:cribo-ad-highlight}. Autonomous driving scenes contain a variety of object sizes and often have long-tailed distributions (e.g., motorcycles, cyclists, construction vehicles). Additionally, large areas of the image are road or background, making small objects like pedestrians or cyclists easy to miss. First, we incorporate \textit{semantic distribution consistent clustering} to encourage better representation of rare classes such as motorcycles or animals in the learned features, as it is important to autonomous driving systems to correctly understand their surroundings. Second, we introduce \textit{object diversity consistent spatial clustering}, to handle imbalanced and diverse object sizes and counts, ranging from large background areas to numerous small objects such as pedestrians and traffic signs. Finally, we propose a \textit{depth-guided spatial clustering} by incorporating the depth information from \lidar point clouds and find that this further improves learned representations by accounting for 3D scene structure, helping in 3D perception tasks as well as accurate 2D segmentation of occluded objects. We achieve this by directly using the depth information without requiring additional compute for a \lidar encoder, as in multi-modal models.

We show that our improved representations perform well on downstream image-based 2D semantic segmentation and 3D object detection tasks. We demonstrate improved generalization capabilities through transfer learning experiments. We conclude by studying the scaling effect of pre-training on AD data and highlight the untapped potential that can be realized by improving data diversity. 

\section{Related work}

\paragraph{Self-Supervised Pre-Training}

Contrastive learning methods have proven to be highly effective in self-supervised learning, popularized by SimCLR \cite{ssl_simclr}. The usage of negative samples and triplet losses help mitigate the representation collapse issue but require large batch sizes, which can be managed with a memory bank \cite{he2020momentum}. Some approaches address this by avoiding explicit negative samples through asymmetry, such as additional predictors, stop-gradients, or momentum encoders \cite{byol, simsiam}. Clustering-based methods further prevent trivial solutions by regularizing sample assignments across clusters \cite{ssl_pcl, sela, deep_cluster, swav, dino}. 
Another set of methods extend self-supervised pre-training to learn representations better suited for dense downstream tasks. These methods use the dense spatial features in their formulation. 
Local-level approaches contrast dense features \cite{wang2021dense, o2020unsupervised, xie2021propagate, cribo, vicreg_local, ibot}, while object-level approaches promote similarity between semantically coherent feature groups \cite{cho2021picie, henaff2021efficient, xie2021propagate, wen2022self, stegmuller2023croc}. 

\cribo \cite{cribo}, builds upon DINO \cite{dino} and CRoC \cite{croc}, to enhance dense visual representation learning by employing object-level nearest neighbor bootstrapping. \cribo showed state-of-the-art performance in various segmentation tasks. 
While these pre-training methods often show good performance when trained on large-scale curated datasets such as ImageNet \cite{dataset_imagenet} or COCO \cite{dataset_coco}, their applicability to  more challenging autonomous driving datasets is currently under explored, with the exception of the older work, MultiSiam \cite{multisiam}. ImageNet pre-trained models are used as initialization in autonomous driving but this transfer of models involves a domain shift and performs worse on tasks like traffic sign classification, which heavily deviate from ImageNet \cite{pretraining_data_ssl}. With this motivation, it is of interest to study and develop pre-training methods for autonomous driving that can generalize between different driving datasets. In this work, we investigate the recent advances achieved by \cribo and propose \mname - scene semantics and structure guided clustering self-supervised pre-training for autonomous driving data.

\begin{figure*}[h!]
    \centering
    \includegraphics[width=.8\textwidth, trim=0 0.1cm 17cm 5.5cm, clip]{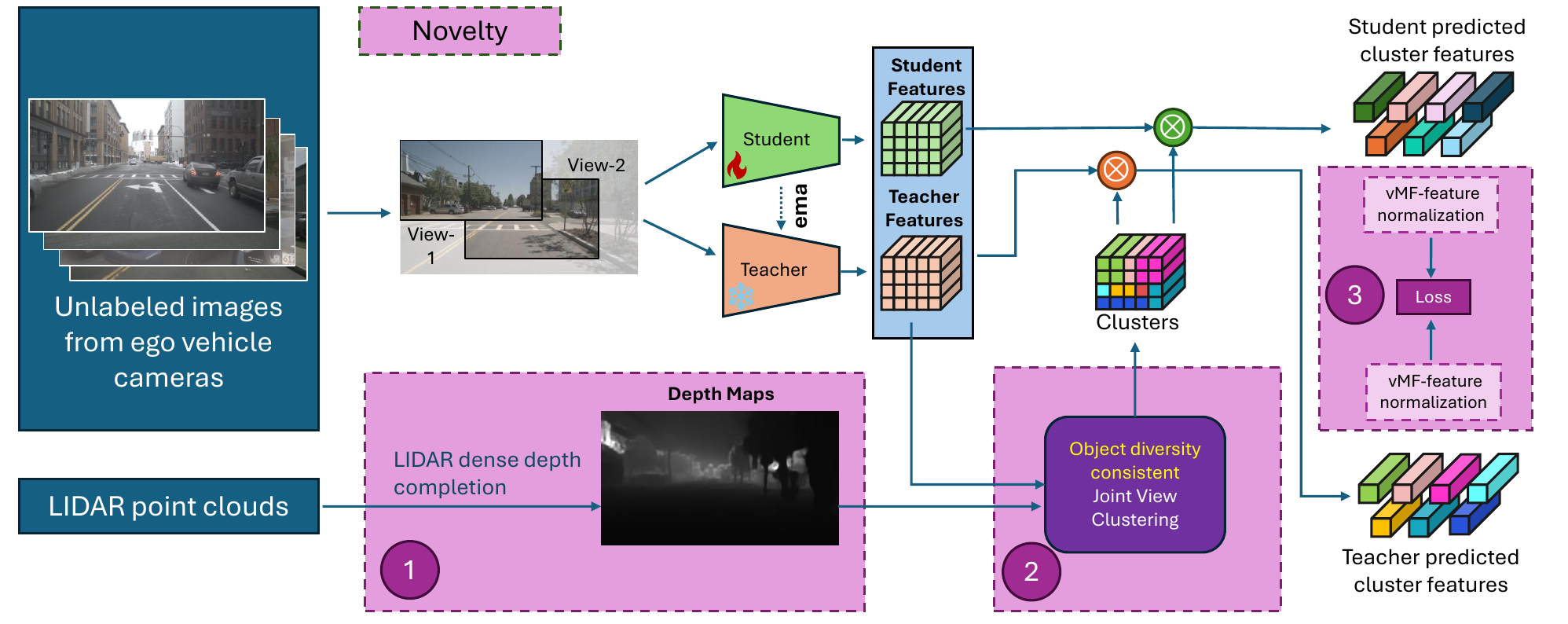}
    \caption{Overview of S3PT. For each image, two different views are extracted and fed into teacher and student networks (same architecture, we use ViT, however, any network can be used). Based on teacher network features and depth information from \lidar, a object-diversity consistent joint-view clustering is performed to extract object-level features. Finally, the student is trained using a vMF-normalized loss formulation which enables flexible and non-uniform semantic distributions.}
    \label{fig:flowchart}
\end{figure*}
\paragraph{Self-Supervised Pre-Training in Autonomous Driving}

Self-supervised learning in autonomous driving often focuses on either LiDAR~\cite{wozniak2024uada3d,chang2024cmda,tsai2023ms3d++,lentsch2024union,Zhang_2023_CVPR,Baur2024ECCV,Sautier_3DV24} or camera~\cite{Zhang_2023_CVPR,zhou2023unidistill,chen2024m}, with fewer studies integrating both~\cite{wozniak2023towards,liu2023stereodistill,liu2024segment,ge2023metabev,puy2024three, sun2023calico}. Li et al.~\cite{li2022simipu} introduced a method with a spatial perception module and a feature interaction module, aligning features using contrastive loss. BEVDistill~\cite{chen2022bevdistill} improved this approach with dense and sparse feature distillation, utilizing a  bird's-eye view (BEV) feature plane for consistent representation. CALICO~\cite{sun2023calico}, an unsupervised solution, employs multi-stage contrastive losses and region-aware distillation to align BEV features from LiDAR and camera. 
Methods using both camera and LiDAR often rely on pre-trained image backbones. Our method, trained solely on autonomous driving data, excels despite the lack of diversity in datasets. It incorporates LiDAR or depth-predicted cues during learning and is entirely unsupervised. Our approach ensures robust camera features that work independently of LiDAR during inference, unlike many fusion methods that heavily depend on LiDAR data during both training and inference. We integrate LiDAR depth information directly into the model without needing a separate LiDAR encoder. This focus on unsupervised pre-training allows us to address the unique challenges of autonomous driving, both in research and in consumer vehicles, which rarely have LiDAR.

\paragraph{Handling Imbalance in Self-Supervised Learning}

Self-supervised methods are usually pre-trained on well-curated datasets such as ImageNet \cite{dataset_imagenet} which contains object-centric images (one predominant object per image) and contains a uniform distribution over the object classes. The uniform distribution of visual concepts is particularly important for several methods. Contrastive learning methods such as SimCLR \cite{ssl_simclr} and MoCo \cite{moco, moco_v2} attempt to spread the features uniformly over the latent space \cite{understanding_contrastive_ssl}. Clustering-based methods such as DeepCluster \cite{deep_cluster}, SwAV \cite{swav}, DINO \cite{dino} and MSN \cite{msn} encourage a uniform distribution over the clusters or pseudo-classes. This is shown to have a negative impact when pre-training on long-tailed datasets \cite{pmsn}. Using temperature schedules \cite{temperature_schedules} and explicit matching to non-uniform prior distributions, PMSN \cite{pmsn} are proposed to address this issue. Recent work indicates that using a clustering formulation based on the von Mises-Fisher distribution can learn reasonably from long-tailed data distributions \cite{on_ppc}, generally performing even better than PMSN. KoLeo-regularization of the prototypes is shown to further improve performance when pre-training on long-tailed data \cite{on_ppc}. While recent methods such as \cribo \cite{cribo} have demonstrated excellent performance when pre-training from scratch on scene data, they still use ImageNet and COCO \cite{dataset_coco} datasets, which have more uniform distributions of objects.

\section{Method}

The DINO-family of methods \cite{dino, ibot, msn, esvit, dino_vmf} use the pretext task of assigning object-centric images to $K$ latent classes. 
Given a dataset $\mathcal{D}$, consider an image, $I \in \mathcal{D}$. 
Consider an encoder model with parameters $\btheta$, that produces an $\normltwo$-normalized global \texttt{[CLS]} (class token) \cite{vit} representation $\z = g_{\btheta}(I)$ such that $\z \in \mathbb{R}^D$ and $\Vert \z \Vert = 1$. The probability of assigning an image to a latent class $k$ under the assumption of a uniform latent class prior is formulated using a softmax operation: $P_k(\z) = \probpr(y=k | \z) = \frac{\exp \left( \langle \W_k, \z \rangle / \tau \right) }{\sum_{j=1}^K \exp \left( \langle \W_j, \z \rangle / \tau \right) }$, where $\W_k \in \mathbb{R}^D$ denotes the weights of the last linear layer, also known as \textit{prototype vectors} and $\tau$ is the temperature. We will refer to this as \textit{semantic clustering}. 
The teacher produces the targets and the student is trained to match the outputs of the teacher. Then, the teacher weights are updated as an exponential moving average (ema) of the student weights.

While DINO only used the global representation in its formulation (appropriate when each image contains a single object), \cribo \cite{cribo} extends it to scene data, by considering each scene as a set of objects. The goal is to learn a representation \( 
\{ \Bar{\boldsymbol{z}}, \boldsymbol{z} \} = f_{\theta}(I) \). Here, $\Bar{\boldsymbol{z}} \in \mathbb{R}^d$ is the global representation and dense representation $\boldsymbol{z} \in \mathbb{R}^{H \times W \times d}$ with downsampled spatial dimensions $H$ and $W$. \cribo identifies the objects in a scene through a \textit{spatial clustering} of the dense spatial features into $M$ clusters to get object-level representations, $\boldsymbol{o}_k \in \mathbb{R}^d$, obtained by mean pooling the spatial features within each cluster. These global and object-level representations are assigned to $K$ different semantic clusters, to obtain probability distributions similar to DINO. The \cribo loss objective is obtained as a combination of three DINO-based loss terms:
\begin{equation}
    \mathcal{L}_{\mathrm{CrIBo}} = \mathcal{L}^g_{\mathrm{cv}} + \mathcal{L}^o_{\mathrm{cv}} + \mathcal{L}^o_{\mathrm{ci}}
\end{equation}
\noindent Each of these three loss terms are formulated as a cross-entropy loss between the teacher and student probability distributions, $H(P^{(t)}, P^{(s)})$. 
The three terms $\mathcal{L}^g_{\mathrm{cv}}$, $\mathcal{L}^o_{\mathrm{cv}}$ and $\mathcal{L}^o_{\mathrm{ci}}$ correspond to global cross-view, object-level cross-view and object-level cross-image consistencies respectively. 

Pre-training with \cribo off-the-shelf on an autonomous driving dataset produces poor representations due to the inappropriate uniformity assumptions made regarding the data distribution over the semantic clusters and area distribution over the spatial clusters. As illustrated in ~\cref{fig:flowchart}, we propose a novel scene semantics and structure guided clustering (S3PT). The semantic clustering is improved by using a vMF normalized formulation \cite{dino_vmf} to improve the pre-training on long-tailed distributions over objects. The spatial clustering is improved to enable diverse object sizes by relaxing the cluster uniformity assumption and using a depth-guided clustering that incorporates the depth information of scenes to further enhance the spatial clustering of objects (see \cref{fig:cribo-vs-sgc-details} for details). We discuss each of these contributions in detail below.

\subsection{Semantic distribution consistent clustering}

\begin{figure}[t!]
    \centering
    \includegraphics[width=.8\linewidth]{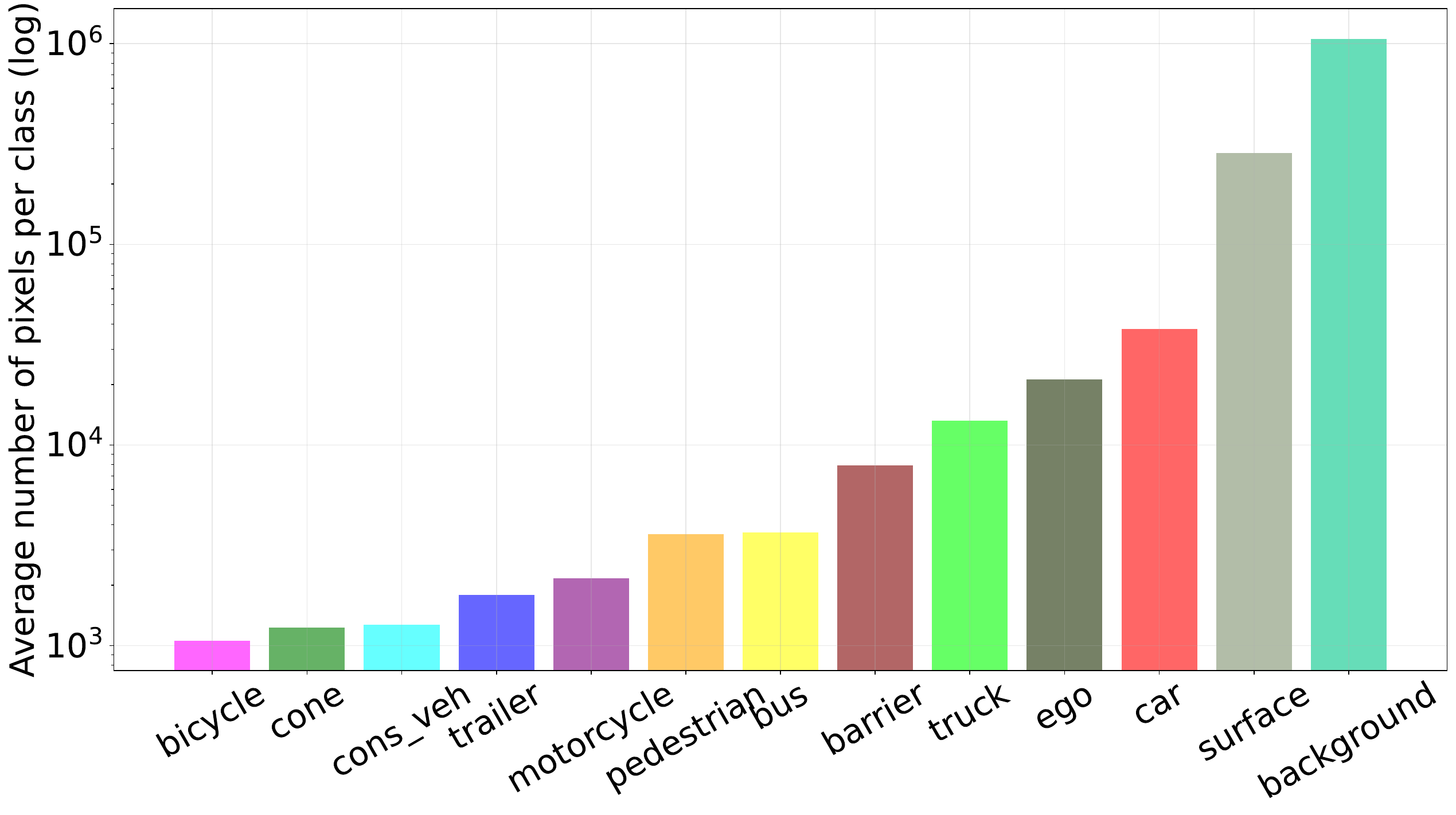}
    \caption{Average number of pixels per class in nuImages dataset.}
    \label{fig:logdistribution}
\end{figure}
Autonomous driving (AD) data typically comprise of a long-tail distribution over the different objects, where some objects such as cars are frequent whereas other objects such as bicycles or motorcycles are relatively rare, as shown in~\cref{fig:logdistribution}. DINO-based methods assume and encourage a uniform distribution over the $K$ semantic clusters \cite{pmsn}, which matches object distributions in pre-training datasets such as ImageNet \cite{dataset_imagenet} or COCO \cite{dataset_coco}. However, this poses a challenge when pre-training on AD datasets. Using vMF normalized formulation of DINO enables more flexible cluster spread \cite{dino_vmf} and is shown to perform well when learning from such long-tailed datasets \cite{on_ppc}. This formulation is obtained by using unnormalized prototypes and modifying the probabilities over the $K$ pseudo-classes as follows:
\begin{equation}
\label{eq: dino_vmf_distribution}
    P_k(\z) = \frac{C(\Vert \W_k \Vert / \tau) \exp \left( \langle \W_k, \z \rangle / \tau \right) }{\sum_{j=1}^K C(\Vert \W_j \Vert / \tau) \exp \left( \langle \W_j, \z \rangle / \tau \right) } .
\end{equation}
\noindent and applying the centering operation in the probability space instead of the logit space as in DINO. Here, $C(\cdot)$ refers to the normalization constant of the vMF distribution. This enhances the semantic clustering by being consistent to the observed long-tail distribution \cite{on_ppc}. We investigate the benefit of using this formulation in Section~\ref{sec:expt_ablation_params}.

\subsection{Scene Distribution consistent clustering}

\begin{figure}
    \centering
    \includegraphics[width=.78\linewidth, trim=2.1cm 0.1cm 32cm 0.1cm, clip]{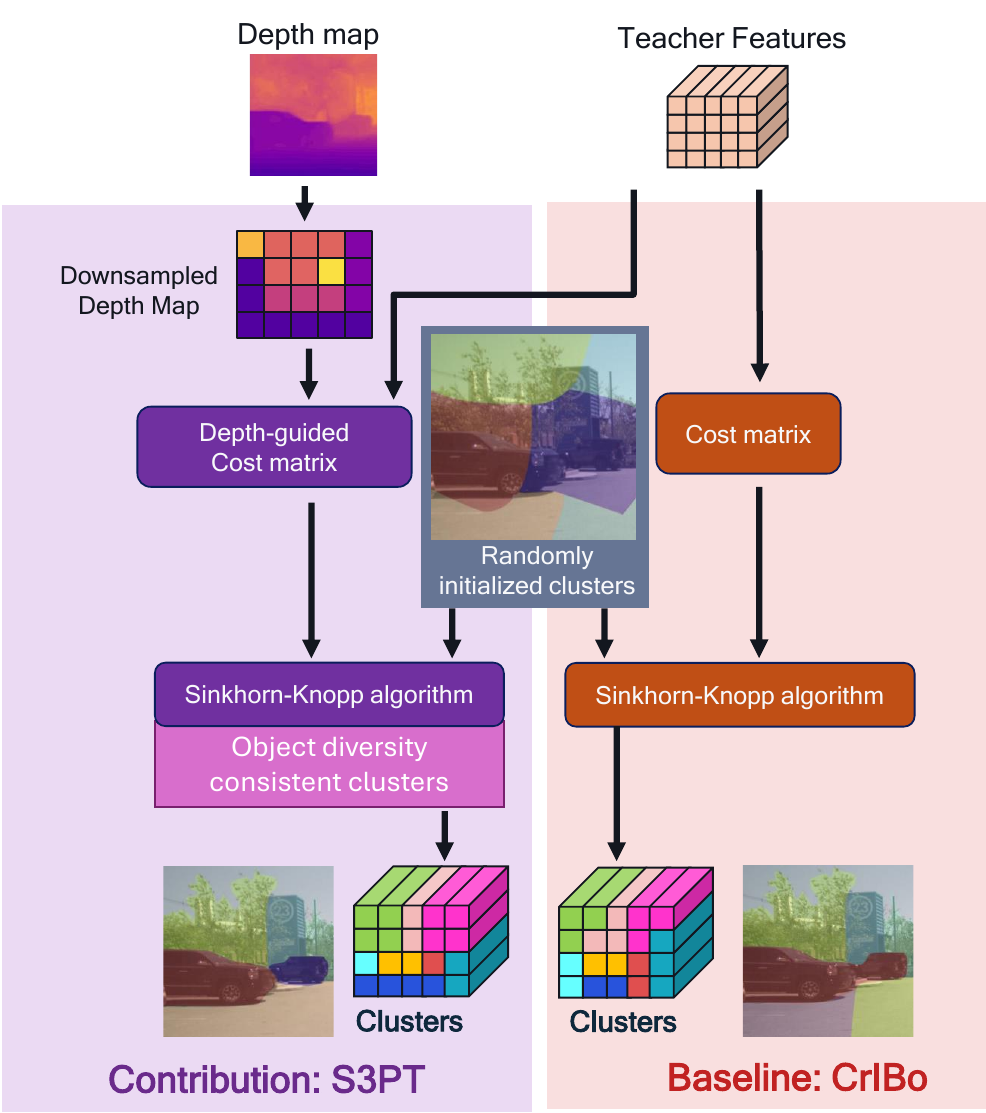}
    \caption{A depth-guided clustering is proposed by using the depth maps to modify the cost matrix. The spatial clustering is modified to use a larger number of clusters and relaxing the uniformity assumption in the Sinkhorn-Knopp algorithm, to enable identification of objects of diverse sizes in a scene.}
    \label{fig:cribo-vs-sgc-details}
\end{figure}
\cribo \cite{cribo} performs spatial clustering of dense features to identify the object regions. This is done by computing a cost matrix based on the dense teacher features. Then, the clustering problem is solved as an optimal transport problem by iteratively running the Sinkhorn-Knopp (SK) optimization algorithm \cite{sinkhorn_cuturi} for multiple iterations. Running the algorithm for multiple iterations strongly encourages a uniform distribution of areas over the spatial clusters. This makes the implicit assumption that all the objects should be approximately equally sized. This is not a meaningful assumption for autonomous driving scenes which feature objects of diverse sizes in a scene - roads and buildings are large regions whereas several objects of interest such as pedestrians, bicycles, traffic cones and animals are small in size. Also, we observe more objects per image in autonomous driving compared to other datasets. To this end, we propose to modify the spatial clustering to remain consistent to the object diversity observed in AD scenes. Firstly, we propose to use a much larger number of spatial clusters compared to \cribo and then, we relax the strong uniformity assumption by limiting the number of SK iterations to just one iteration. We demonstrate in Section~\ref{sec:expt_ablation_params} that this enhances the spatial clustering through empirical analysis and improves performance on rarer objects.

\subsection{Depth-guided spatial clustering}
\label{sec: method_depth_aided_clustering}

The spatial clustering is formulated as a joint-view clustering of $2P$ spatial tokens from the two views of an image into $M$ clusters. This is done by solving an optimal transportation problem. Consider the transportation cost matrix $\boldsymbol{T} \in \mathbb{R}^{2P \times M}$. We modify this transportation cost to also account for the depth information, available through the \lidar data as: $\boldsymbol{T} = \boldsymbol{T}^{\mathrm{(CrIBo)}} + \beta \boldsymbol{T}^{\mathrm{(depth)}}$. The depth-based cost corresponding to token $i$ and cluster centroid $j$, with depths $d_i$ and $d_j$ respectively, is computed as $\boldsymbol{T}_{ij}^{\mathrm{(depth)}} = \Vert d_i - d_j \Vert_2$. Since the \lidar point clouds are sparse, we consider a depth completion approach to obtain depth information at all image pixels. This enables the spatial clustering to account for the scene structure made apparent through depth guidance.

\section{Experiments}
\label{sec: experiments}

\begin{table*}[ht!]
\centering
\small
\begin{tabular}{lccccll}
\toprule
Note & \makecell{Num.\\clusters} & vMF & SK iters. & \makecell{Depth\\weight} & \makecell{Linear\\(mIoU)} & \makecell{Mask Trans.\\(mIoU)} \\
\midrule
\cribo baseline & 32 & No & 5 & -- & 20.50 & 41.67 \\
 \multirowcell{2}[0pt][l]{+ semantic distribution \\ \xspace\xspace\xspace consistent clustering } & \cellcolor{white!60!lightgray} 32 & \cellcolor{white!60!lightgray} Yes & \cellcolor{white!60!lightgray} 5 & \cellcolor{white!60!lightgray} -- & \cellcolor{white!60!lightgray} \textbf{26.07} \textsuperscript{\textcolor{red}{(+5.57)}} & \cellcolor{white!60!lightgray} \textbf{45.04} \textsuperscript{\textcolor{red}{(+3.37)}} \\
 & & & & & & \\
 \midrule
 \multirowcell{7}[0pt][l]{+ Object diversity \\ \xspace\xspace\xspace consistent \\ \xspace\xspace\xspace spatial clustering}& 8 & Yes & 5 & -- & 26.13 & 43.40 \\
 & 16 & Yes & 5 & -- & 26.18 & 43.87 \\
 & 32 & Yes & 5 & -- & 26.07 & 45.04 \\
 & 64 & Yes & 5 & -- & 24.86 & 45.85 \\
 & 128 & Yes & 5 & -- & 24.15 & 46.01 \\
 & 32 & Yes & 1 & -- & 25.24 & 45.47 \\
 & \cellcolor{white!60!lightgray} 128 & \cellcolor{white!60!lightgray} Yes & \cellcolor{white!60!lightgray} 1 & \cellcolor{white!60!lightgray} -- & \cellcolor{white!60!lightgray} \textbf{28.10} \textsuperscript{\textcolor{red}{(+2.03)}} & \cellcolor{white!60!lightgray} \textbf{47.46} \textsuperscript{\textcolor{red}{(+2.42)}} \\
 \midrule
 \multirowcell{3}[0pt][l]{+ Depth-guided\\ \xspace\xspace\xspace spatial clustering \\ } & 128 & Yes & 1 & 0.5 & 36.80 & 51.10 \\
   & \cellcolor{white!60!lightgray} 128 & \cellcolor{white!60!lightgray} Yes & \cellcolor{white!60!lightgray} 1 & \cellcolor{white!60!lightgray} 1.0 & \cellcolor{white!60!lightgray} 37.65 & \cellcolor{white!60!lightgray} \textbf{51.51} \textsuperscript{\textcolor{red}{(+4.05)}} \\
   & \cellcolor{white!60!lightgray} 128 & \cellcolor{white!60!lightgray} Yes & \cellcolor{white!60!lightgray} 1 & \cellcolor{white!60!lightgray} 4.0 & \cellcolor{white!60!lightgray} \textbf{38.06} \textsuperscript{\textcolor{red}{(+9.96)}} & \cellcolor{white!60!lightgray} 51.04 \\
\bottomrule
\end{tabular}
\caption{Test results on nuImages using a frozen backbone and different heads - Mask Transformer and Linear probing with different configurations trained for \textbf{100 epochs}. We show how different components of our method contribute to the performance. We can see that each component brings large improvements compared to the \cribo baseline and our best model significantly outperforms the baseline.}
\label{tab:ablation_results}
\end{table*}

\begin{figure*}[ht]
    \centering
    \includegraphics[width=.94\textwidth]{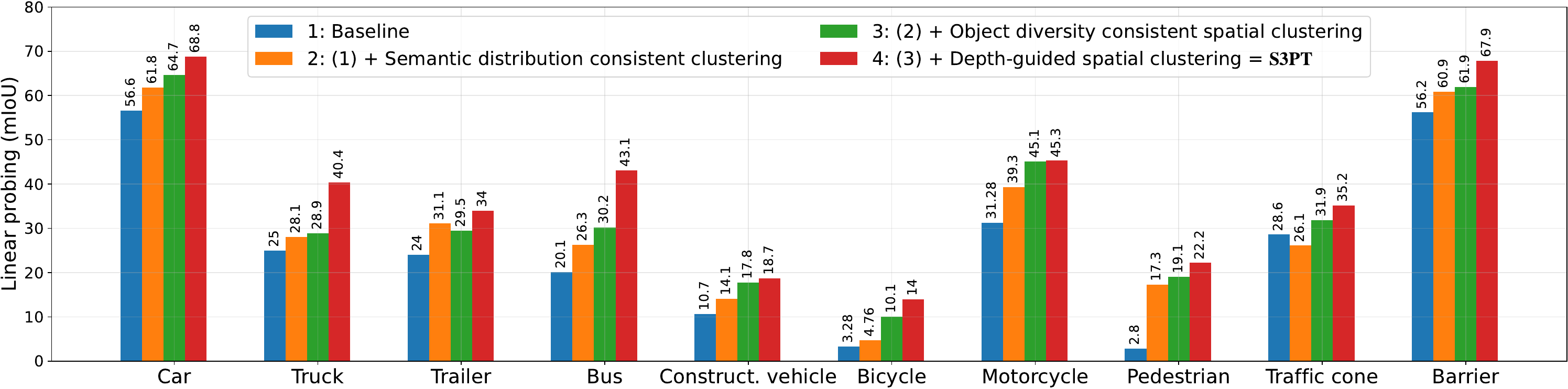}
    \caption{Object-wise segmentation performance of models from Tab.~\ref{tab:ablation_results} using the Mask Transformer head. The models are obtained by sequentially applying the proposed modifications to \cribo baseline to finally achieve \textbf{\mname}. Our proposed modifications improve performance of all classes and particularly improved performance is observed on underrepresented classes like bus and construction vehicle and also small-sized objects  as bicycle, motorcycle, pedestrians and traffic cones.}
    \label{fig:object_wise_performance}
\end{figure*}

\subsection{Ablation Studies}
\label{sec:expt_ablation_params}

We start with the default configuration of the \cribo baseline and we pre-train a ViT-Small/16 model for 100 epochs. Then, we sequentially add or adapt different components to the pre-training method. We evaluate the learned representations on semantic segmentation of nuImages by freezing the backbone model and training a Segmenter \cite{segmenter} head - either a linear probing layer or a more complicated Mask Transformer for 160K iterations (see supplementary for detailed evaluation configuration). The detailed results for each of the following configurations are reported in \cref{tab:ablation_results} and the object-specific performances are shown in \cref{fig:object_wise_performance}. 

\paragraph{Semantic distribution consistent clustering}
We investigate the impact of using the vMF normalized formulation of \cribo and find that it improves the overall performance. While the baseline could not identify less frequent objects at all, the vMF normalized version shows a promising improvement (see \cref{fig:object_wise_performance}). 

\paragraph{Scene Distribution consistent clustering} 
We find that simply changing the number of clusters has little impact on the performance, because of the assumption of uniform cluster/object sizes. However, if the cluster uniformity is relaxed by reducing the SK iterations to 1, we observe that using a larger number of clusters is able to learn diverse representations accounting for smaller and underrepresented objects. 
This significantly improves performance on smaller objects as bicycles, motorcycles and traffic cones in \cref{fig:object_wise_performance}. Relaxing the cluster uniformity while still using a small number of clusters is not beneficial as this could still merge several smaller objects together to achieve fewer clusters.

\begin{table*}[ht]
\centering
\small
\resizebox{\linewidth}{!}{
\begin{tabular}{lcccccc}
\toprule
\textbf{Model} & \textbf{Dataset (size)} & \textbf{Epochs} & \multicolumn{2}{c}{\textbf{nuImages}} & \multicolumn{2}{c}{\textbf{Cityscapes}} \\
\cmidrule(lr){4-5} \cmidrule(lr){6-7}
 & & & \textbf{Mask Transformer} & \textbf{Linear Probing} & \textbf{Mask Transformer} & \textbf{Linear Probing} \\
\midrule
\multicolumn{7}{l}{\textit{ViT-Small/16}} \\
\textbf{DINO}~\cite{dino} & nuScenes (240K)  & 500 & \textbf{58.19} & 35.27 &  53.70 & 28.75  \\
\textbf{\cribo}~\cite{cribo}& nuScenes (240K) & 500 & 50.31 & 37.94 & 44.95 & 30.30 \\
\textbf{S3PT (Ours)} & nuScenes (240K) & 500 & 55.70 & \textbf{42.13} & \textbf{55.41 } & \textbf{37.32}\\
\midrule

\multicolumn{7}{l}{\textit{ViT-Base/16}} \\
\textbf{DINO}~\cite{dino} & nuScenes (240K)  & 100 & \textbf{49.65} & 37.01 & 37.64 & 26.84 \\
\textbf{\cribo}~\cite{cribo} & nuScenes (240K) & 100 & 42.45 & 31.82 &29.82 & 29.73 \\
\textbf{S3PT (Ours)} & nuScenes (240K) & 100 & 48.92 & \textbf{42.94} & \textbf{45.41} & \textbf{32.62} \\
\bottomrule
\end{tabular}
}
\caption{Test results of models pre-trained in self-supervised manner on nuScenes and evaluated on nuImages and Cityscapes using Mask Transformer head and Linear probing. Models are trained in an unsupervised way for \textbf{500 (ViT-S)} and \textbf{100 (ViT-B)} epochs, and then used as a frozen backbone with segmentation head (Linear probing or Mask Transformer), trained for 160k iterations. }
\label{tab:full_training_results}
\end{table*}
\begin{figure*}[h]
    \centering
   
    \centering
    \includegraphics[width=.88\textwidth]{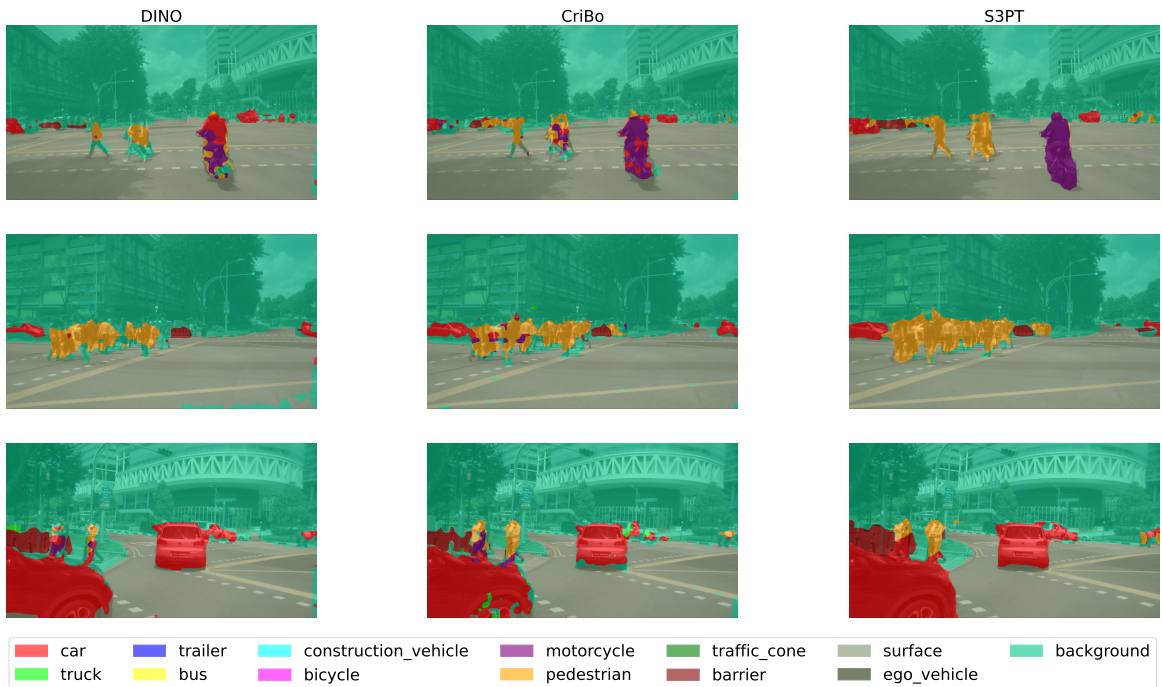}
     \caption{Comparison of qualitative results on the semantic segmentation task on nuImages with linear probing. Our method (\textbf{S3PT}) excels, especially on small and long-tail classes like motorcycles and pedestrians. Despite the majority of the image being background or surface, our method accurately segments rare classes and distant objects, unlike other methods.}
    \label{fig:results_visualization}
\end{figure*}

\paragraph{Depth-guided spatial clustering}
In order to obtain the depth information at every token location from the sparse raw \lidar point clouds, we use the fast IP-Basic depth interpolation \cite{ip_basic}. We ablate between different weights $\beta$ for the depth term. We find that including this depth for clustering is beneficial in improving the model performance (see \cref{tab:ablation_results}). This is because the depth signal helps in incorporating a naturally occurring depth bias to separate objects present at different depths into distinct clusters, even if they are visually similar, as illustrated in \cref{fig:cribo-ad-highlight}.

\subsection{Experimental Results}
\label{sec: expt_ablation_depth}

We conducted extensive experiments with longer training and different models (ViT-S/16 and ViT-B/16) to benchmark our proposed method with competitive baselines DINO \cite{dino} and \cribo \cite{cribo}. \cribo is current SOTA method for SSL pre-training on complex scenes and an improvement of CroC \cite{croc}, while DINO is one of the most popular image-based SSL approaches. All the models in this section are pre-trained in a fully self-supervised manner on the nuScenes \cite{caesar2020nuscenes}. ViT-S/16 models undergo 500 epochs of training, while ViT-B/16 models are trained for 100 epochs only due to computational constraints (see supplementary for full training setup). Our method, S3PT, not only performs well on the original datasets but also excels in domain adaptation and 3D object detection tasks, indicating that our method learns generalized and robust features, making it a versatile solution for various computer vision tasks.

\paragraph{Semantic Segmentation} 
In \cref{tab:full_training_results}, we compare our models  with DINO and \cribo baseline methods.  We freeze the backbone model and evaluate the learned features on the semantic segmentation downstream task using nuImages \cite{caesar2020nuscenes} and Cityscapes \cite{cordts2016cityscapes} datasets.
We train the Mask Transformer and Linear probing heads for 160K iterations similar to \cribo \cite{cribo} (see supplementary for complete evaluation protocol).
This setup allows us to assess the effectiveness of our method in learning robust features that generalize well across different datasets. With all the datasets and heads, we significantly improve over \cribo, which is our main baseline. Interestingly, \cribo outperforms DINO in linear probing but significantly underperforms DINO when using a more complex Mask Transformer head. We hypothesize that DINO features are more suitable for more complex segmentation models and are not as generic as S3PT. On nuImages, our model with the Mask Transformer head achieves mIoU lower than that of DINO, 55.70 vs. 58.19, but still significantly improving over \cribo (50.31). However, in linear probing, our model significantly outperformed DINO, scoring 42.13 compared to 35.27, indicating superior raw feature quality. On the Cityscapes dataset, our model excelled, achieving the highest scores with both Mask Transformer head (55.41) and linear probing (37.32). This demonstrates our model’s robustness and ability to generalize across different domains (we further explore this in \cref{tab:domainadaptation}). Furthermore, in \cref{fig:results_visualization} we can observe that in comparison to other methods, our approach significantly improves the segmentation of long-tail classes and distant objects (motorcycles or pedestrians), even though the majority of the image consists of background or surface. It also excels in segmenting distant or occluded objects, thanks to the enhanced spatial features.

\begin{table}[ht!]
\centering
\small
\resizebox{0.5\textwidth}{!}{
\begin{tabular}{lcccc}
\toprule
\multirow{2}{*}{\textbf{Model}} & \multirow{2}{*}{\textbf{Dataset}} & \multirow{2}{*}{\textbf{Iterations}} & \textbf{Mask} & \textbf{Linear} \\
& & & \textbf{Transformer} & \textbf{Probing} \\

\midrule
\multicolumn{5}{l}{\textit{ViT-Small/16}} \\
\textbf{DINO}~\cite{dino} & nuI $\rightarrow$ CS  & 160k & 17.43 & 19.83 \\
\textbf{\cribo}~\cite{cribo}& nuI $\rightarrow$ CS & 160k & 16.91 & 18.61 \\
\textbf{S3PT (Ours)} & nuI $\rightarrow$ CS & 160k & \textbf{18.39} & \textbf{22.57} \\
\midrule

\multicolumn{5}{l}{\textit{ViT-Base/16}} \\
\textbf{DINO}~\cite{dino} & nuI $\rightarrow$ CS  & 160k & 11.07 & 14.60 \\
\textbf{\cribo}~\cite{cribo} & nuI $\rightarrow$ CS & 160k  & 12.29 & 15.27 \\
\textbf{S3PT (Ours)} & nuI $\rightarrow$ CS & 160k & \textbf{18.36} & \textbf{16.65} \\

\midrule
\multicolumn{5}{l}{\textit{ViT-Small/16}} \\
\textbf{DINO}~\cite{dino} & CS $\rightarrow$ nuI  & 160k & 21.15 & 27.40 \\
\textbf{\cribo}~\cite{cribo} & CS $\rightarrow$ nuI & 160k & 19.07 & 23.82 \\
\textbf{S3PT (Ours)} & CS $\rightarrow$ nuI & 160k & \textbf{24.19} & \textbf{30.89} \\
\midrule

\multicolumn{5}{l}{\textit{ViT-Base/16}} \\
\textbf{DINO}~\cite{dino} & CS $\rightarrow$ nuI  & 160k & 19.73 &22.62 \\
\textbf{\cribo}~\cite{cribo} & CS $\rightarrow$ nuI & 160k & 17.07 & 17.83
 \\
\textbf{S3PT (Ours)} & CS $\rightarrow$ nuI & 160k & \textbf{21.96} & \textbf{27.97} \\

\bottomrule
\end{tabular}
}
\caption{Test results on domain transfer between nuImages (\textbf{nuI}) and Cityscapes (\textbf{CS}) using Mask Transformer Head and Linear Probing. ViT-S models are trained for 500 epochs and ViT-B models for 100 epochs in an unsupervised manner on nuScenes. These models are then used as a frozen backbone for downstream segmentation task, trained for 160k iterations.}
\label{tab:domainadaptation}
\end{table}

\paragraph{Domain Generalization} To further test the generalization capabilities of our models, we perform domain transfer experiments in \cref{tab:domainadaptation} between nuImages and Cityscapes. We use frozen backbone (trained with S3PT or baseline method) with segmentation head. Segmentation heads are trained on nuImages, then tested on Cityscapes and vice versa. S3PT demonstrates superior performance in these domain adaptation tasks, indicating that it learns more generic and effective features compared to other methods. Following the results we can observe that S3PT learns better spatial features than DINO or \cribo and achieves higher performance across all metrics.

\paragraph{3D Object Detection} Finally, in \cref{tab:3dod_results} we evaluate the performance of our method on 3D object detection tasks. 
We use the frozen pre-trained backbones and fine-tune the PETR detection head, reporting the mean Average Precision (mAP) and nuScenes Detection Score (NDS). This comparison highlights the effectiveness of our method in learning transferable features that perform well across different tasks, including 3D object detection. 
Unlike other methods, such as DINO and \cribo, our method considers the depth information that enables an improved understanding of the 3D scene geometry. This enables our method to learn representations that perform well not only in 2D segmentation tasks, but also in 3D object detection.
Our model demonstrates superior domain generalization capabilities due to its ability to effectively learn spatial and object-specific features. 
This enhanced learning capability allows our model to perform better across different autonomous driving domains, making it a more robust and versatile solution for various applications.

\begin{table}[t!]
\centering
\small
\begin{tabular}{lccc}
\toprule
\textbf{Model} & \textbf{Method} & \textbf{mAP} & \textbf{NDS} \\
\midrule

{\textbf{PETR} (V2-99)}\textsuperscript{$\dagger$} & Supervised & 36.5 & 43.1 \\
\midrule

\multirowcell{3}{\textbf{PETR} \\ (ViT-Small/16)}
& DINO~\cite{dino} & 16.45 & 22.99 \\
 & \cribo~\cite{cribo}& 13.08 & 21.76 \\
& S3PT (Ours) & \textbf{18.17} & \textbf{24.74} \\
\midrule

\multirowcell{3}{\textbf{PETR} \\ (ViT-Base/16)} 
 & DINO~\cite{dino} & 17.23 & 24.14\\
& \cribo~\cite{cribo}&  13.05 & 22.02 \\
& S3PT (Ours) & \textbf{19.24} & \textbf{26.49} \\
 \bottomrule
 \multicolumn{4}{l}{\textsuperscript{$\dagger$} Based on public checkpoint from PETR \cite{petr}} \\

\end{tabular}
\caption{Comparison on how backbones pre-trained by DINO, \cribo and S3PT (ours) perform on 3D object detection tasks. We use a frozen pre-trained backbone and fine tune the PETR detection head and report mAP and NDS over nuScenes dataset.}
\label{tab:3dod_results}
\end{table}

\begin{figure}[t!]
    \centering
    \includegraphics[width=0.7\linewidth]{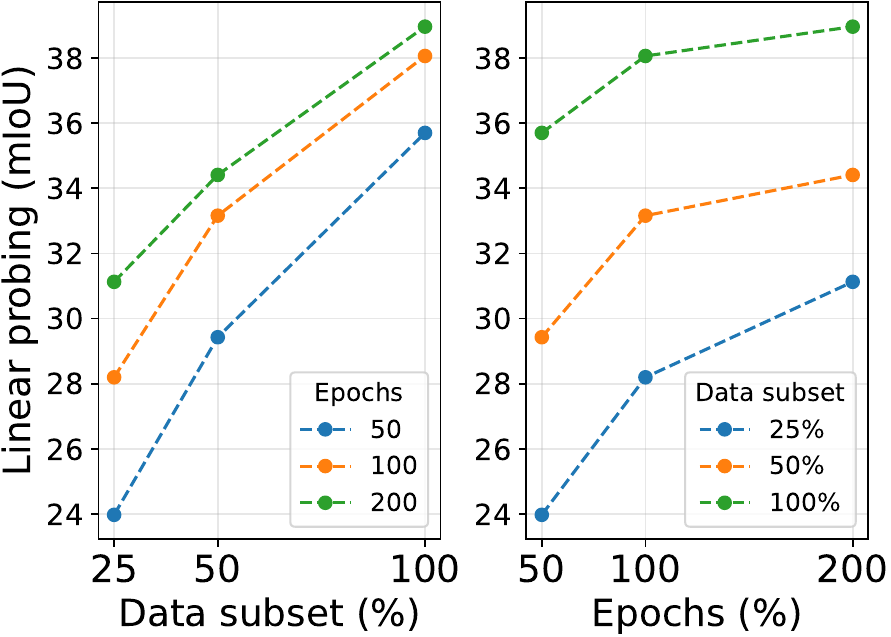}
    \caption{Segmentation performance (using linear probing) of S3PT models pre-trained on different subsets of the nuScenes dataset and for varying number of epochs. While performance saturates with longer trainings, there is potential improve performance by scaling the pre-training dataset. }
    \label{fig:data_subset_analysis}
\end{figure}

\subsection{Study on data diversity and training length}
\label{sec: study_data_diversity}

We evaluate the importance of pre-training length (in terms of number of epochs) and the size of the pre-training datasets. We consider smaller subsets of the nuScenes dataset by randomly sampling $25\%$ and $50\%$ of the \textit{sequences} and using all the frames from all cameras in the selected sequences. The linear probing results are reported in Fig.~\ref{fig:data_subset_analysis}. Although longer trainings are beneficial, we find that larger and more diverse datasets yield higher performance gains. While performance begins to plateau with extended training (right subfigure), there remains potential for improvement through the use of larger, more varied datasets (left subfigure, we can see that performance has not saturated yet).
This can enable pre-training on autonomous driving datasets to not only improve further but also become the superior de-facto choice over using Imagenet pre-trained models for AD tasks.

\section{Conclusions}
\label{sec: discussion}

We introduce \mname, a novel scene semantics and structure guided clustering to improve clustering-based self-supervised learning methods pre-trained on scene data. Through extensive evaluation on a wide range of downstream tasks that depend on different types of information, we demonstrate that our proposed method learns effective feature representations that generalize across various tasks and autonomous driving datasets. Our experiments show improved representation of rarer objects on segmentation and 3D object detection tasks. Finally, we highlight the potential to significantly improve the performance by scaling the pre-training dataset size. We also could expect that in combination with domain adaptation method, S3PT would generalize even better. While this study focuses on autonomous driving applications, the results in terms of domain generalization and pre-training design choices can also have high impact on other applications domains.
\paragraph{Acknowledgment}
This work was partially supported by the Wallenberg AI, Autonomous Systems and Software Program (WASP) funded by the Knut and Alice Wallenberg Foundation.

{\small
\bibliographystyle{ieee_fullname}
\bibliography{references}
}

\newpage

\section*{Supplementary Materials}
\appendix

\section{Configuration details}

\subsection{Pre-training setup}
\label{sec:pertrain}
Here, we provide the complete pre-training configuration for all the methods that we evaluate in \cref{sec: expt_ablation_depth} of the main paper. The DINO and \cribo methods are pre-trained using their publicly available repositories and we follow their default configuration. For completeness, we provide the full configuration details for DINO in \cref{table:app_hyperparams_dino} and for \cribo in \cref{table:app_hyperparams_cribo}. Our proposed S3PT is based on the \cribo repository and we use a similar configuration to the defaults, except for our proposed modifications. The full configuration details of S3PT is provided in \cref{table:app_hyperparams_s3pt}.

\begin{table}[h]
  \centering
  \scriptsize
  \begin{tabular}{lll}
    \toprule
    Hyper-parameter & ViT-Small/16 & ViT-Base/16 \\
    \midrule
    training epochs & $500$ & $100$ \\
    batch size & $256$ & $256$ \\
    learning rate & $5\sce{-4}$ & $7.5\sce{-4}$ \\
    warmup epochs & $10$ & $10$ \\
    freeze last layer epochs & $1$ & $3$ \\
    min. learning rate & $1\sce{-5}$ & $2\sce{-6}$ \\
    weight decay & $0.04 \rightarrow 0.4$ & $0.04 \rightarrow 0.4$ \\
    stochastic depth & $0.1$ & $0.1$ \\
    gradient clip & - & $0.3$ \\
    optimizer & adamw & adamw \\
    fp16 & \cmark & \cmark \\
    \midrule
    momentum & $0.996 \rightarrow 1.0$ & $0.996 \rightarrow 1.0$ \\
    global crops & $2$ & $2$ \\
    global crops scale & $[0.25, 1.0]$ & $[0.25, 1.0]$ \\
    local crops & $10$ & $10$ \\
    local crops scale & $[0.05, 0.25]$ & $[0.05, 0.25]$ \\
    \midrule
    head mlp layers & $3$ & $3$ \\
    head hidden dim. & $2048$ & $2048$ \\
    head bottleneck dim. & $256$ & $256$ \\
    norm last layer & \xmark & \cmark \\
    num. prototypes & $65536$ & $65536$ \\
    \midrule
    teacher temp. & $0.04 \rightarrow 0.07$ & $0.04 \rightarrow 0.07$ \\
    temp. warmup epochs & $30$ & $50$ \\
    student temp. & $0.1$ & $0.1$ \\
    \bottomrule
  \end{tabular}
  \caption{Hyperparameter settings for DINO}
  \label{table:app_hyperparams_dino}
\end{table}

\begin{table}[h]
  \centering
  \scriptsize
  \begin{tabular}{lll}
    \toprule
    Hyper-parameter & ViT-Small/16 & ViT-Base/16 \\
    \midrule
    training epochs & $500$ & $100$ \\
    batch size & $256$ & $256$ \\
    learning rate & $5\sce{-4}$ & $7.5\sce{-4}$ \\
    warmup epochs & $10$ & $10$ \\
    freeze last layer epochs & $1$ & $3$ \\
    min. learning rate & $1\sce{-5}$ & $2\sce{-6}$ \\
    weight decay & $0.04 \rightarrow 0.4$ & $0.04 \rightarrow 0.4$ \\
    stochastic depth & $0.1$ & $0.1$ \\
    gradient clip & - & $0.3$ \\
    optimizer & adamw & adamw \\
    fp16 & \cmark & \cmark \\
    \midrule
    momentum & $0.996 \rightarrow 1.0$ & $0.996 \rightarrow 1.0$ \\
    global crops & $2$ & $2$ \\
    global crops scale & $[0.25, 1.0]$ & $[0.32, 1.0]$ \\
    \midrule
    head mlp layers & $3$ & $3$ \\
    head hidden dim. & $2048$ & $2048$ \\
    head bottleneck dim. & $256$ & $256$ \\
    norm last layer & \xmark & \cmark \\
    num. prototypes & $65536$ & $65536$ \\
    \midrule
    teacher temp. & $0.04 \rightarrow 0.07$ & $0.04 \rightarrow 0.07$ \\
    temp. warmup epochs & $30$ & $50$ \\
    student temp. & $0.1$ & $0.1$ \\
    \midrule
    sinkhorn lambda & $20.0$ & $20.0$ \\
    sinkhorn iterations & $5$ & $5$ \\
    pos alpha & $[1.0, 1.0]$ & $[1.0, 1.0]$ \\
    which features & last & last \\
    num spatial clusters & $32$ & $32$ \\
    queue size & $25000$ & $25000$ \\
    \bottomrule
  \end{tabular}
  \caption{Hyperparameter settings for \cribo}
  \label{table:app_hyperparams_cribo}
\end{table}

\begin{table}[h]
  \centering
  \scriptsize
  \begin{tabular}{lll}
    \toprule
    Hyper-parameter & ViT-Small/16 & ViT-Base/16 \\
    \midrule
    training epochs & $500$ & $100$ \\
    batch size & $256$ & $256$ \\
    learning rate & $5\sce{-4}$ & $7.5\sce{-4}$ \\
    warmup epochs & $10$ & $10$ \\
    freeze last layer epochs & $1$ & $3$ \\
    min. learning rate & $1\sce{-5}$ & $2\sce{-6}$ \\
    weight decay & $0.04 \rightarrow 0.4$ & $0.04 \rightarrow 0.4$ \\
    stochastic depth & $0.1$ & $0.1$ \\
    gradient clip & - & $0.3$ \\
    optimizer & adamw & adamw \\
    fp16 & \cmark & \cmark \\
    \midrule
    momentum & $0.996 \rightarrow 1.0$ & $0.996 \rightarrow 1.0$ \\
    global crops & $2$ & $2$ \\
    global crops scale & $[0.25, 1.0]$ & $[0.32, 1.0]$ \\
    \midrule
    head mlp layers & $3$ & $3$ \\
    head hidden dim. & $2048$ & $2048$ \\
    head bottleneck dim. & $256$ & $256$ \\
    norm last layer & \xmark & \xmark \\
    num. prototypes & $65536$ & $65536$ \\
    vmf normalization & \cmark & \cmark \\
    centering & probability & probability \\
    \midrule
    teacher temp. & $0.04 \rightarrow 0.07$ & $0.04 \rightarrow 0.07$ \\
    temp. warmup epochs & $30$ & $50$ \\
    student temp. & $0.1$ & $0.1$ \\
    \midrule
    sinkhorn lambda & $20.0$ & $20.0$ \\
    sinkhorn iterations & $1$ & $1$ \\
    pos alpha & $[1.0, 1.0]$ & $[1.0, 1.0]$ \\
    depth alpha & $[4.0, 4.0]$ & $[4.0, 4.0]$ \\
    which features & last & last \\
    num spatial clusters & $128$ & $128$ \\
    queue size & $2500$ & $2500$ \\
    \bottomrule
  \end{tabular}
  \caption{Hyperparameter settings for S3PT}
  \label{table:app_hyperparams_s3pt}
\end{table}

\subsection{Evaluation protocols}

In this section, we provide details about the evaluation protocols and configurations used for different downstream evaluation experiments, namely semantic segmentation, domain generalization and 3D object detection. 

\subsubsection{Semantic segmentation}

We obtain dense feature representations from frozen backbones and only train the head network for semantic segmentation on nuImages \cite{caesar2020nuscenes} and Cityscapes \cite{cordts2016cityscapes} datasets. We consider a similar evaluation protocol as in \cribo \cite{cribo} and use the linear decoder head from Segmenter \cite{segmenter}. This linear probing head maps the token features ($16\times 16$ patches) to class assignments and then uses a bilinear upsampling to transform the outputs to image pixel dimensions. We additionally, also evaluate the Mask Transformer decoder head, with the same default configuration proposed in Segmenter. We use the \texttt{mmsegmentation} 1.2.2 \cite{mmsegmentation}  library for evaluation and use the default 160K iterations training schedule. For each pre-training method and dataset, the results reported in the tables are the best results after considering the following set of learning rates: \texttt{\{8e-4, 3e-4, 8e-5\}}, similar to other works which perform such evaluations.
For nuImages, we use the same dataset configuration setup as the publicly available \texttt{ade20k} dataset configuration in \texttt{mmsegmentation}. For Cityscapes, we use the \texttt{cityscapes\_768x768} configuration. The configurations for the linear probing and Mask Transformer decoder heads in Segmenter are available in \texttt{mmsegmentation}.
For domain generalization experiments, we use the same semantic segmentation models (frozen backbone and decoder head) trained using the above described setup.

\subsubsection{3D object detection}
For our test and training for 3D object detection task we use \texttt{mmdetection3D} 1.4.0~\cite{mmdet3d2020}. For benchmarking, we use camera-only 3D object detector PETR~\cite{petr} (author's implementation in mmdetection3D library\footnote{\url{https://github.com/open-mmlab/mmdetection3d/tree/main/projects/PETR}}). In short, PETR encodes the position information of 3D coordinates into image features, creating 3D position-aware features. This allows object queries to perceive these features and perform end-to-end object detection. Essentially, PETR transforms multi-view images into a unified 3D space by combining positional information directly with the image features. This approach enables the model to detect objects in 3D space using only camera data, without relying on LiDAR or other sensors. 

The original backbone for PETR in \texttt{mmdetection3D} implementation is VoVNetCP, based on the VoVNet~\cite{lee2019energy}, is designed for efficient and effective feature extraction. It uses a unique One-Shot Aggregation (OSA) module, which concatenates features from multiple layers only once, reducing computational overhead and improving efficiency.

For original benchmark, denoted as \textit{supervised} in \cref{tab:3dod_results} (of the main paper), we used the backbone weights provided by the authors\footnote{\url{https://drive.google.com/file/d/1ABI5BoQCkCkP4B0pO5KBJ3Ni0tei0gZi/view}}. This model is pre-trained on DDAD15M and then, trained on nuScenes train set in a supervised manner.

For self-supervised pre-training of image backbones with methods: DINO, \cribo and S3PT (ours) we used weights obtained in pre-training, described in \cref{sec: expt_ablation_depth} of the main paper (detailed above in \cref{sec:pertrain}). Next, we trained PETR together with these backbones, however, with frozen image backbone weights, thus gradients does not flow through to the image backbone. Please, refer to \cref{table:app_petr} for detailed training setting and we use the \texttt{cyclic-20e} scheduler from \texttt{mmdetection3D}.

\begin{table}[h]
  \centering
  \scriptsize
  \begin{tabular}{ll}
  \toprule
    Hyper-parameter & PETR \\
    \midrule
    img size & $(320, 800)$ \\
    grid size & $[512,512,1]$ \\
    voxel size  & $[0.2, 0.2, 8]$ \\
    \midrule

    training epochs & $20$ \\
    batch size & $2$ \\
    learning rate & $3\sce{-4}$ \\
    warmup epochs & $1$ \\
    weight decay & $0.01$  \\
    stochastic depth & $0.1$ \\
    gradient clip & $35$ \\
    optimizer & Adamw \\
    fp16 & \xmark \\
    momentum & $0.85 \rightarrow 1.0$ \\
    global crops & $2$ \\
    \midrule
    \textbf{PETR head} & \\
    head hidden dim. & $384/784 (ViT-S/B)$ \\
    num query & $900$ \\
    num layers & $6$ \\ 
    num heads & $8$ \\ 
    feedforward channels & $2048$ \\
    \bottomrule
  \end{tabular}
  \caption{Hyperparameter settings for 3D object detection}
  \label{table:app_petr}
\end{table}

\section{Additional results}

\subsection{ViT backbones with different patch sizes}

\begin{figure*}[ht]
  \centering
    \includegraphics[width=0.95\linewidth]{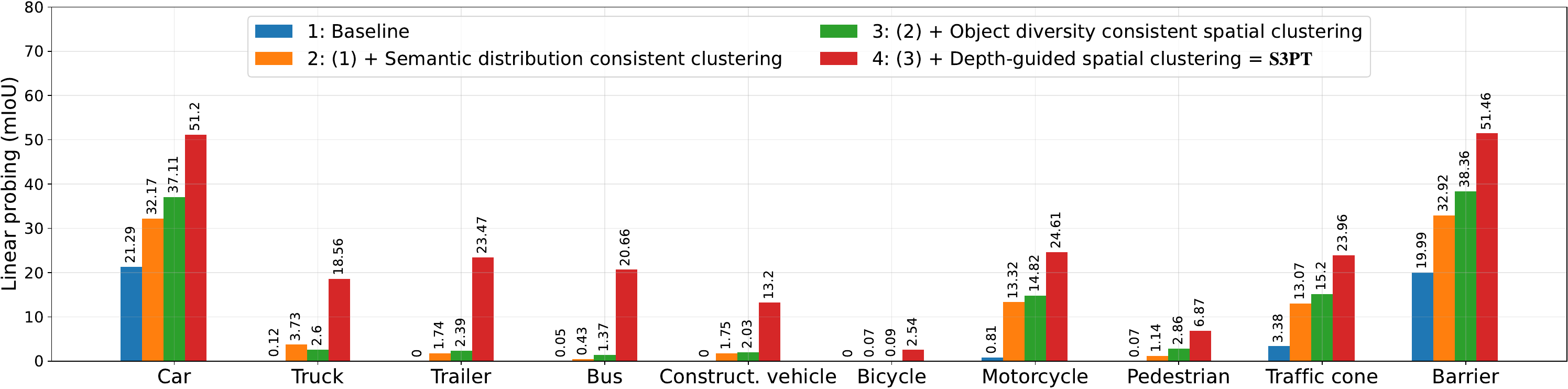}
    \caption{Object-wise segmentation performance of models from \cref{tab:ablation_results} using the linear probing head. The models are obtained by sequentially applying the proposed modifications to CrIBo baseline to finally achieve S3PT.}
    \label{fig:linear_object_wise_perf}
\end{figure*}

In addition to the standard ViTs with patch size 16 evaluated in the main paper, we also evaluate the impact of using different patch sizes in ViT-Small model. Specifically, we evaluate patch sizes $14$, $16$ and $32$. We avoid patch size $8$, as it is extremely compute expensive to train. We pre-train these models using S3PT for 100 epochs with the same pre-training configuration. Only difference is that we reduce the number of spatial clusters to $32$ for ViT-Small/32 due to the decreased number of patch tokens. We evaluate the models on nuImages semantic segmentation task with a Mask Transformer head and report the results in \cref{table:patch_size_s3pt}. We find smaller patch sizes to perform better, which is in agreement with other self-supervised pre-training methods. This is especially beneficial for dense prediction tasks, where smaller patch sizes enable more granular predictions.

\begin{table}[h]
  \centering
  \scriptsize
\begin{tabular}{lccc}
\toprule
\textbf{Class} & \multicolumn{3}{c}{\textbf{mIoU}} \\
\cmidrule{2-4}
& ViT-Small/14  &  ViT-Small/16  &  ViT-Small/32 \\
\midrule
Car & 68.57 & 68.47 & 63.59 \\
\hline
Truck  & 36.1 & 41.52 & 30.72 \\
\hline
Trailer & 33.43 & 30.13 & 24.83 \\
\hline
Bus & 42.59 & 34.9 & 32.12 \\
\hline
Construction Vehicle & 26.21 & 14.55 & 15.37 \\
\hline
Bicycle & 12.47 & 9.33 & 10.5 \\
\hline
Motorcycle & 48.18 & 47.09 & 42.65 \\
\hline
Pedestrian & 23.41 & 22.1 & 14.75 \\
\hline
Traffic Cone & 33.54 & 35.96 & 25.83 \\
\hline
Barrier & 65.82 & 66.82 & 61.74 \\
\hline
\textbf{Overall} & \textbf{50.97} & \textbf{50.04} & \textbf{46.04} \\
\hline
\end{tabular}
  \caption{Semantic segmentation performance on nuImages (with Mask Transformer head) of ViT-Small models with different patch sizes, after pre-training with S3PT on nuScenes dataset}
  \label{table:patch_size_s3pt}
\end{table}

\subsection{Object-wise performance of our contributions}

In \cref{fig:object_wise_performance} of the main paper, we showed the object-wise performance of sequentially adding our contributions on the nuImages semantic segmentation task with a Mask Transformer head. We show the linear probing performance of adding our contributions in \cref{fig:linear_object_wise_perf} above.

\subsection{Average distance to different objects in the dataset }
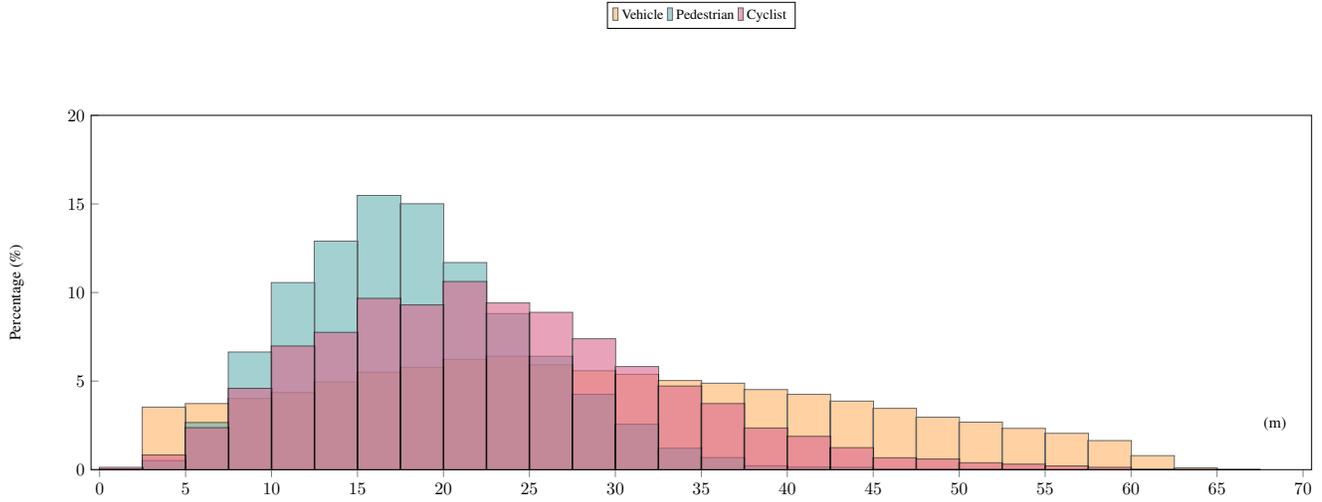
\begin{figure*}[h!]
    \centering
    \begin{tikzpicture}
\begin{axis} [ybar = .05cm,
    bar width = 25pt, 
    width=1.5\textwidth,
    height=0.5\textwidth,
    tick pos=left,
   xmin = 0, 
    xmax = 70, 
    ymin=0,
    ymax=20,
    enlarge x limits = {abs=0.5},
    tick label style={font=\normalsize},
    legend style={font=\footnotesize},
    label style={font=\small},
    y label style={at={(axis description cs:0.0,.5)},anchor=south},
    x label style={at={(axis description cs:0.97,0.24)},anchor=north},
     legend style={legend columns=-1,at={(0.5,1.32)},anchor=north},
    legend image code/.code={
        \draw [#1] (0cm,-0.1cm) rectangle (0.1cm,0.15cm);},
    bar shift = 0.0,
    xlabel = (m),
    ylabel = Percentage (\%)
]
\addplot [fill=orange!60,opacity=0.6] coordinates{
(1.25, 0.05) (3.75, 3.53) (6.25, 3.73) (8.75, 4.02) (11.25, 4.35) (13.75, 4.95) (16.25, 5.50) (18.75, 5.78) (21.25, 6.23) (23.75, 6.40) (26.25, 5.92) (28.75, 5.58) (31.25, 5.38) (33.75, 5.03) (36.25, 4.88) (38.75, 4.52) (41.25, 4.25) (43.75, 3.87) (46.25, 3.46) (48.75, 2.96) (51.25, 2.68) (53.75, 2.33) (56.25, 2.05) (58.75, 1.64) (61.25, 0.79) (63.75, 0.10) (66.25, 0.00)
};

\addplot [fill=teal!60,opacity=0.6] coordinates{
(1.25, 0.03) (3.75, 0.51) (6.25, 2.66) (8.75, 6.64) (11.25, 10.56) (13.75, 12.90) (16.25, 15.48) (18.75, 15.01) (21.25, 11.69) (23.75, 8.81) (26.25, 6.40) (28.75, 4.25) (31.25, 2.56) (33.75, 1.21) (36.25, 0.68) (38.75, 0.22) (41.25, 0.14) (43.75, 0.13) (46.25, 0.04) (48.75, 0.02) (51.25, 0.03) (53.75, 0.01) (56.25, 0.00) (58.75, 0.00) (61.25, 0.00) (63.75, 0.00) (66.25, 0.00)
};

\addplot [fill=purple!60,opacity=0.6] coordinates{
(1.25, 0.13) (3.75, 0.83) (6.25, 2.37) (8.75, 4.59) (11.25, 6.98) (13.75, 7.75) (16.25, 9.67) (18.75, 9.30) (21.25, 10.62) (23.75, 9.41) (26.25, 8.88) (28.75, 7.39) (31.25, 5.82) (33.75, 4.72) (36.25, 3.73) (38.75, 2.35) (41.25, 1.88) (43.75, 1.24) (46.25, 0.67) (48.75, 0.60) (51.25, 0.39) (53.75, 0.31) (56.25, 0.21) (58.75, 0.13) (61.25, 0.01) (63.75, 0.00) (66.25, 0.00)
};

\legend {Vehicle, Pedestrian, Cyclist};
 
\end{axis} 
\end{tikzpicture}

    \caption{Average distance to different objects in the dataset cap at 65 meters}
    \label{fig:distance-to-ob}
\end{figure*}

In~\cref{fig:distance-to-ob} can see that on average most of the objects are between 15-25m away from from the ego vehicle. However, some objects are far away, which makes them harder to detect. In \cref{tab:3ddistance} we show that adding more and more distant objects, substantially decrease the performance of PETR. Nevertheless, We observe that PETR with S3PT (our backbone) does not drop as much in performance as other models which features have not learnt any 3D cues. Additionally it performs better on long-tailed distribution objects. 

\begin{table*}[h!]
\centering
\caption{Performance on 3D object detection with different backbones, depending on the distance/range. We can see that our method is more robust towards far away objects (note, we remove construction vehicle class).}
\label{tab:3ddistance}
\begin{tabular}{lccgccgccg}
\toprule
\textbf{Class Names} & \multicolumn{3}{c}{\textbf{0-20m}} & \multicolumn{3}{c}{\textbf{0-30m}} & \multicolumn{3}{c}{\textbf{0-60m}} \\
\cmidrule(lr){2-4} \cmidrule(lr){5-7} \cmidrule(lr){8-10} 
& \textbf{Dino} & \textbf{Cribo} & \textbf{Ours} & \textbf{Dino} & \textbf{Cribo} & \textbf{Ours} & \textbf{Dino} & \textbf{Cribo} & \textbf{Ours} \\
\midrule
car & 0.515 & 0.489 & 0.555 & 0.288 & 0.333 & 0.396 & 0.276 & 0.246 & 0.299 \\
truck & 0.300 & 0.344 & 0.368 & 0.159 & 0.229 & 0.281 & 0.166 & 0.151 & 0.195\\
bus & 0.276 & 0.237 & 0.427 & 0.172 & 0.139 & 0.312 & 0.077 & 0.056 & 0.135 \\
trailer & 0.153 & 0.103 & 0.222 & 0.144 & 0.082 & 0.168 & 0.122 & 0.053 & 0.168 \\
pedestrian & 0.455 & 0.401 & 0.477 & 0.337 & 0.269 & 0.349 & 0.226 & 0.205 & 0.272 \\
motorcycle & 0.357 & 0.331 & 0.405 & 0.295 & 0.210 & 0.307 & 0.197 & 0.196 & 0.275  \\
bicycle & 0.033 & 0.02 & 0.014 & 0.016 & 0.027 & 0.028 & 0.006 & 0.005 & 0.022 \\
traffic cone & 0.352 & 0.335 & 0.370 & 0.285 & 0.287 & 0.297 & 0.263 & 0.263 & 0.285  \\
barrier & 0.592 & 0.452 & 0.421 & 0.418 & 0.465 & 0.430 & 0.369 & 0.343 & 0.368 \\
mAP & \textbf{0.337} & \textbf{0.299} & \textbf{0.362}& \textbf{0.261} &\textbf{ 0.227} & \textbf{0.285} & \textbf{0.189} & \textbf{0.169} &\textbf{ 0.224} \\
\bottomrule
\end{tabular}
\end{table*}

\subsection{Additional qualitative results}

\begin{figure*}[t!]
    \centering
    \includegraphics[width=\textwidth]{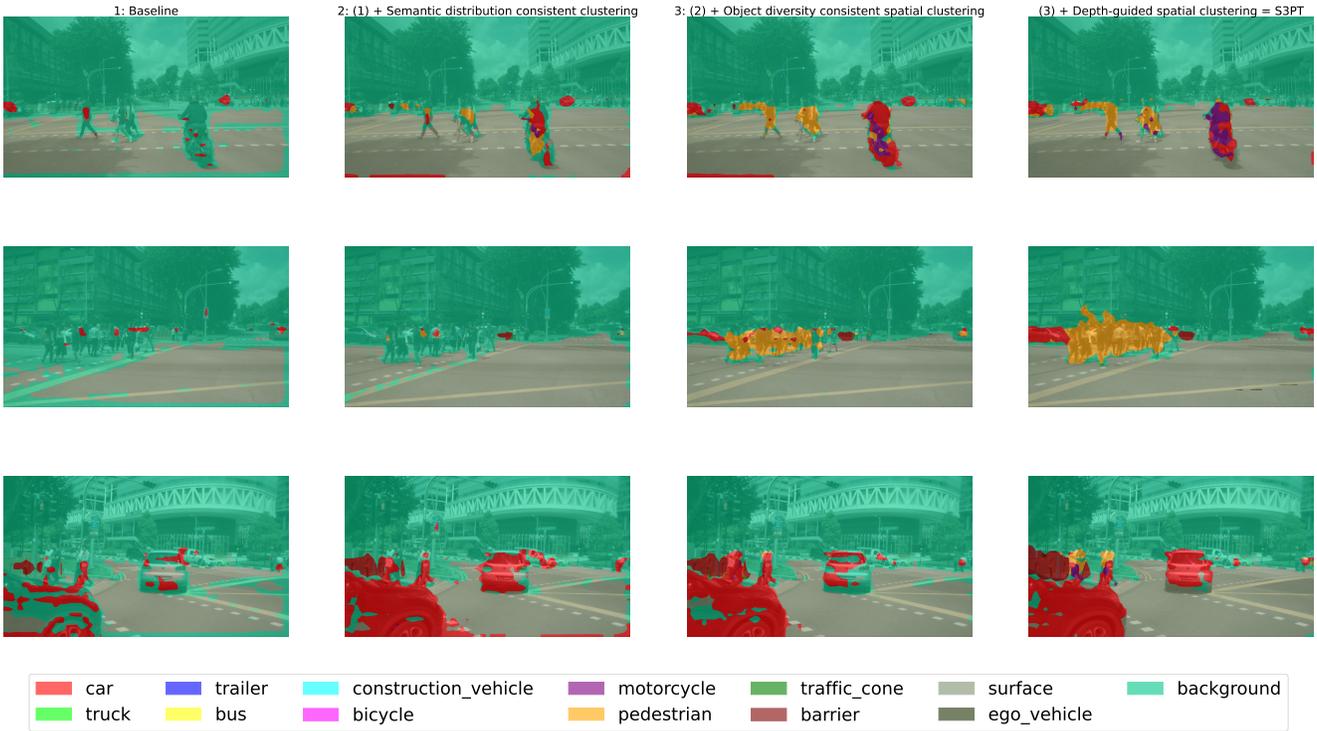}
    \caption{Qualitative results obtained by adding each of our proposed components to the baseline \cribo method. Note that these are visualizations corresponding to the ablation experiments shown in Tab. 1 and the object-wise performances shown in Fig. 5, where we pre-trained a ViT-S/16 backbone for only 100 epochs and then trained a linear probing Segmenter head. With a longer pre-training for 500 epochs using S3PT, we demonstrate improved further segmentation quality (see Fig. 6 and Tab. 2).}
    \label{fig:addresults}
\end{figure*}

In \cref{fig:addresults} we present additioanl results for models used in hyper parameter search in Table 1 and Figure 5. We can observe how adding each component to our training schema influences final segmentation quality results. Note, these presented models are ViT-S pretrained with only 100 epochs with Linear Probing, we used this shorter trainig for hyperparameter search. Full results require 500 epochs for ViT-S and 100 for ViT-B.

\end{document}